\documentclass[11pt,letterpaper]{article}
\usepackage{ijcnlp2017}
\usepackage{times}
\usepackage{latexsym}
\usepackage{booktabs}
\usepackage{subcaption}
\usepackage{graphicx}
\usepackage{color,soul}
\usepackage{amsmath,amssymb}
\usepackage{multicol}
\usepackage{multirow}
\usepackage[normalem]{ulem}
\usepackage{pifont}
\usepackage{enumitem}
\usepackage{diagbox}

\usepackage{color}

\usepackage{url}

\usepackage{lingmacros}

\newcommand{\hh}{\mathbf{h}}

\ijcnlpfinalcopy

\def\ijcnlppaperid{***}

\title{Evaluating Layers of Representation in Neural Machine Translation on Part-of-Speech and Semantic Tagging Tasks}

\author{Yonatan Belinkov$^1$ ~ Llu\'{i}s M\`arquez$^2$ ~  Hassan Sajjad$^2$ ~ \\  \textbf{Nadir Durrani}$^2$  ~ \textbf{Fahim Dalvi}$^2$ ~ \textbf{James Glass}$^1$ \\\\
$^1$MIT Computer Science and Artificial Intelligence Laboratory, Cambridge, MA 02139, USA \\
{\tt \{belinkov, glass\}@mit.edu} \\
$^2$Qatar Computing Research Institute, HBKU, Doha, Qatar  \\
{\tt \{lmarquez, hsajjad, ndurrani, faimaduddin\}@qf.org.qa}
}

\date{}

\begin{document}

\maketitle

\begin{abstract}
 While neural machine translation (NMT) models 
  provide improved translation quality in an elegant, end-to-end 
  framework, it is less clear what they 
  learn about language. 
  Recent work has started evaluating the quality of vector representations learned by NMT models on morphological and syntactic tasks.
  In this paper, we investigate the representations learned at different layers of  NMT encoders.  We train NMT systems on parallel data and use the trained %
  models to extract features for training a classifier on two tasks: part-of-speech and semantic tagging. We then measure the performance of the classifier as a proxy to the quality of the original NMT model for the given task. Our quantitative 
analysis yields interesting insights regarding 
representation learning in  NMT models. For instance, we find that higher layers are better at learning semantics while lower layers  
tend to be 
better for part-of-speech tagging. We also observe little 
effect of the target language on source-side representations, 
especially with higher quality NMT models.\footnote{Our code is available at \url{https://github.com/boknilev/nmt-repr-analysis}.}
\end{abstract}

\section{Introduction}

Neural machine translation (NMT) offers an elegant end-to-end architecture, 
while at the same time improving translation quality.  However, little 
is known about the inner workings of these models and their interpretability is limited. 
Recent work  has 
started exploring what kind of linguistic information such 
models learn
on morphological \cite{vylomova2016word,belinkov:2017:ACL}  and syntactic levels \cite{shi-padhi-knight:2016:EMNLP2016,E17-2060}.

One observation that has been made is that lower layers in the neural MT network learn different kinds of information than higher layers. For example, \newcite{shi-padhi-knight:2016:EMNLP2016} and \newcite{belinkov:2017:ACL} found that representations from lower layers of the NMT encoder are more predictive of word-level linguistic properties like part-of-speech (POS) and morphological tags, whereas higher layer representations are more predictive of more global syntactic information. 
In this work, we take a first step towards understanding what NMT models learn about semantics. We 
evaluate
NMT representations from different layers 
on a semantic tagging task and compare to  
the results on a POS tagging task.  
We believe that understanding semantics learned in NMT can facilitate using 
semantic 
information for improving NMT systems,  
as previously shown for
non-neural MT \cite{P07-1005,C10-1081,W11-1012, 
wu2011lexical,C12-1083,bazrafshan-gildea:2013:Short,bazrafshan-gildea:2014:EMNLP2014}.

For the semantic (SEM) tagging task, 
we use the dataset recently introduced by \newcite{bjerva-plank-bos:2016:COLING}. This is a lexical semantics task: given a sentence, the goal is to assign each word with a tag representing a semantic class. The classes capture nuanced meanings that are ignored in most POS tag schemes. For instance, proximal and distal demonstratives (e.g.\ \textit{this} and \textit{that}) are typically assigned the same POS tag (\texttt{DT}) but receive different SEM tags (\texttt{PRX} and \texttt{DST}, respectively), and proper nouns are assigned different SEM tags depending on their type (e.g.,\ geopolitical entity, organization, person, and location). As another example, consider pronouns like \textit{myself}, \textit{yourself}, and \textit{herself}. They may have reflexive or emphasizing functions, as in (\ref{ex:en-ref}) and (\ref{ex:en-emp}), respectively:
\enumsentence{Sarah bought herself a book\label{ex:en-ref}}
\enumsentence{Sarah herself bought a book\label{ex:en-emp}}
In these examples, \textit{herself} has the same POS tag (\texttt{PRP}) but different SEM tags:  \texttt{REF} for the reflexive function and \texttt{EMP} for the emphasizing function. 

Capturing semantic distinctions of this sort can be important for producing an accurate translation. For instance, example (\ref{ex:en-ref}) would be translated into Spanish with the reflexive pronoun \textit{se}, whereas example (\ref{ex:en-emp}) would be translated with the intensifier \textit{misma}. Therefore, a machine translation system needs to learn different representations of \textit{herself} in the two sentences.

In order to assess the quality of the representations learned by NMT models, 
we adopt the following methodology 
from \newcite{shi-padhi-knight:2016:EMNLP2016} and \newcite{belinkov:2017:ACL}. 
We first train an NMT system on parallel data. Given a sentence, we extract representations from the pre-trained NMT model and train a word-level %
classifier to predict a tag for each word. Our assumption is that the performance of the classifier 
reflects the quality of the representation for the given task.

We compare POS and SEM tagging quality with representations 
extracted from different layers or from models trained on different target languages, while keeping the English source-side fixed.  Our analysis yields interesting insights regarding representation learning in NMT:
\begin{itemize}
\item Consistent with previous work, 
we find that lower layer representations are  usually 
better for POS tagging. However, we also observe that  representations
from higher layers of the NMT 
encoder are better at capturing semantics, even though these are word-level labels. 
This is especially true with tags that are more semantic in nature such as discourse functions and noun concepts.
\item In contrast to previous work, we observe little effect of the target language on source-side representation. We find that the effect of target language diminishes as the size of data used to train the NMT model increases.
\end{itemize}

The rest of the paper is organized as follows: Section \ref{sec:methodology} presents our methodology. Section \ref{sec:exp} describes the data and experimental setup. In Section \ref{sec:results}, we present the results. Section~\ref{sec:related-work} reviews related work and Section \ref{sec:conclusion} concludes the paper.

\section{Methodology}
\label{sec:methodology}
\begin{figure}[t]
	\centering
	\includegraphics[width=\linewidth]{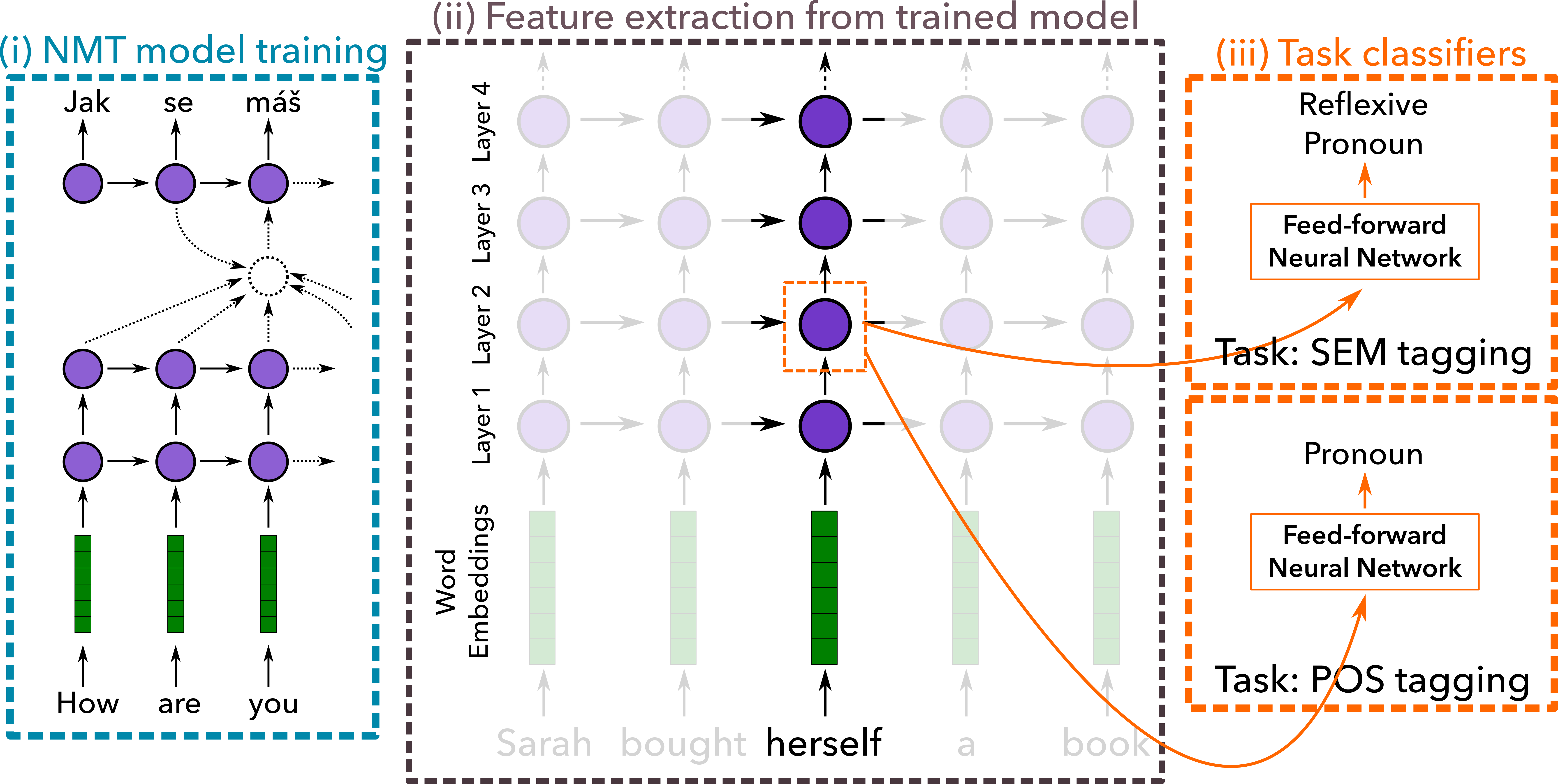}
	\caption{Illustration of our approach, after \cite{belinkov:2017:ACL}: (i) NMT system trained on     parallel data; (ii)  features extracted from pre-trained model;
    (iii) classifier trained using the extracted features. 
       We train classifiers on either SEM or POS  tagging  using features from different hidden layers (here: layer 2). 
    } 
	\label{fig:approach}
\end{figure}

Given a parallel 
corpus of source and target sentence pairs, 
we train an NMT system with a standard sequence-to-sequence model with attention  \cite{bahdanau2014neural,sutskever2014sequence}. 
After training the NMT system, we fix its parameters and treat it as a feature generator for our classification task. 
Let $\hh^k_j $ denote the output of the \mbox{$k$-th} layer of the encoder at the \mbox{$j$-th} word.
Given another corpus of sentences, 
where each word is annotated with a label, we train a classifier that takes   
$\hh^k_j$
as input features and maps words to labels. 
We then measure the performance of the classifier as a way to evaluate the quality of the 
representations
generated by the NMT system. 
By extracting different NMT features we can obtain a quantitative comparison of representation learning 
quality in  the NMT model for the given task. For instance, we may vary $k$ in order to 
evaluate representations learned at different encoding layers.

In our case, we first train NMT systems on parallel corpora of an English source and several target languages. Then we train separate classifiers for predicting POS and SEM tags using the features $\hh^k_j$ that are obtained from the English encoder and evaluate their accuracies. Figure \ref{fig:approach} illustrates the process. 

\section{Data and Experimental Setup}
\label{sec:exp}

\subsection{Data}
\paragraph{MT}
We use the fully-aligned United Nations corpus for training NMT models \cite{ZIEMSKI16.1195}, which includes 11 million
multi-parallel sentences in six languages: Arabic (Ar), Chinese (Zh), English (En), French (Fr), Spanish (Es), and Russian (Ru). We train En-to-* models on 
the first 2 million sentences of the train set, using the official train/dev/test split.
This dataset has the benefit of multiple alignment of the six languages, which allows for comparable cross-linguistic analysis. 

Note that the parallel dataset 
is only used for training the NMT model. The classifier is then trained on the supervised data (described next) and all accuracies are reported on the English test sets.

\paragraph{Semantic tagging}
\newcite{bjerva-plank-bos:2016:COLING} introduced a new 
sequence labeling task, for tagging words with semantic (SEM) tags in context.  
This is a good task to use as a starting point
for investigating semantics
because: \emph{i)} tagging words with semantic labels is very simple, compared to building complex relational semantic structures; \emph{ii)} it provides a large supervised dataset to train on, 
in contrast to most 
available datasets on word sense disambiguation, lexical substitution, and lexical similarity; 
and \emph{iii)} the proposed 
SEM tagging task is an abstraction over 
POS tagging aimed at being language-neutral, and oriented to multi-lingual semantic parsing, all relevant aspects to MT.
We provide here a brief overview of the task and its associated dataset, and refer to \cite{bjerva-plank-bos:2016:COLING,E17-2039} for more details.

The semantic classes abstract over redundant POS distinctions and disambiguate useful cases inside a given POS tag. 
Examples (\ref{ex:en-ref}-\ref{ex:en-emp}) above illustrate how fine-grained semantic distinctions may be important for generating accurate translations.
Other examples of SEM tag distinctions include 
determiners like \textit{every}, \textit{no}, and \textit{some} 
that are typically 
assigned a single POS tag (e.g.\ \texttt{DT} in the Penn Treebank), but have different SEM tags, reflecting universal quantification (\texttt{AND}), negation (\texttt{NOT}), and existential quantification (\texttt{DIS}), respectively.   
The comma, whose POS tag is a punctuation mark, is assigned different SEM tags representing conjunction, disjunction, or apposition, according to its discourse function.  Proximal and distant demonstratives (\textit{this} vs.\ \textit{that}) have different SEM tags but the same POS tag. Named-entities, whose POS tag is usually a single tag for proper nouns, are disambiguated into several classes such as geo-political entity, location, organization, person, and artifact. 
Other nouns are divided into ``role'' entities (e.g. \textit{boxer}) and ``concepts'' (e.g. \textit{wheel}), 
a distinction reflecting existential consistency: an entity can have multiple roles but cannot be two different concepts. 

The dataset annotation scheme 
includes 66 fine-grained tags grouped in 13 coarse categories. We use the silver part of the dataset; see Table \ref{tab:data-semtags} for some statistics.

\paragraph{Part-of-speech tagging}
For POS tagging, we simply use the Penn Treebank with the standard split (parts 2-21/22/23 for train/dev/test); see Table \ref{tab:data-semtags} for statistics. There are 34 POS tags.

\begin{table}[t]
\centering
\begin{tabular}{l|l|rrr}
\toprule
& & Train & Dev & Test \\
\midrule
\multirow{2}{*}{POS} & Sentences & 38K & 1.7K & 2.3K \\
& Tokens & 908K & 40K & 54K \\
\midrule
\multirow{2}{*}{SEM} & Sentences & 42.5K & 6.1K & 12.2K \\
& Tokens & 937.1K & 132.3K & 265.5K \\
\bottomrule
\end{tabular}
\caption{Statistics of the part-of-speech and semantic tagging datasets.}
\label{tab:data-semtags}
\end{table}

\subsection{Experimental Setup}

\paragraph {Neural MT} We use the \texttt{seq2seq-attn} 
toolkit \cite{kim2016} to train 4-layered long short-term memory (LSTM) \cite{hochreiter1997long} attentional encoder-decoder NMT systems with 500 dimensions for both word embeddings and LSTM states. 
We compare both unidirectional and bidirectional encoders and experiment with different numbers of layers. 
Each system is trained with SGD for 20 epochs and the model with the best loss on the development set is used for generating features for the classifier.

\paragraph {Classifier}  

The classifier is modeled as a feed-forward neural network with one hidden layer, dropout (ratio of 0.5), a ReLU activation function, and a softmax layer onto the label set size. 
The hidden layer is of the same size as the input coming from the NMT system (i.e.\ 500 dimensions). 
The classifier has no explicit access to context other than the hidden representation generated by the NMT system, which allows us to focus on the quality of the representation.
We chose this simple formulation as our goal is not to improve the state-of-the-art on the supervised task, but rather to analyze the quality of the NMT representation for the task.\footnote{Previous work found that a non-linear classifier leads to 
relative behavior similar 
to a linear classifier but overall better results~\cite{qian-qiu-huang:2016:P16-11,belinkov:2017:ACL}.}
We train the classifier 
for 30 epochs by minimizing the cross-entropy loss using Adam \cite{kingma2014adam} with default settings.  
Again, we use the model with the best loss on the development set for evaluation. 

\paragraph{Baselines and an upper bound}

we consider two
baselines: most frequent tag (MFT) for each  
word according to the training set
(with the global majority tag for unseen words); and unsupervised word embeddings (UnsupEmb) as features for the classifier, which shows what a simple task-independent distributed representation can achieve. For the unsupervised word embeddings, we train a Skip-gram negative sampling model \cite{mikolov2013distributed} with 500 dimensional vectors on the English side of the parallel data, to mirror the NMT word embedding size.  
We also report an upper bound of  directly 
training an encoder-decoder
on word-tag sequences (Word2Tag), simulating what an NMT-style model can 
achieve by directly optimizing for 
the tagging tasks.

\section{Results}
\label{sec:results}

\begin{table}[t]
\centering
\begin{tabular}{l|ccc}
\toprule
& MFT & UnsupEmb &  Word2Tag \\
\midrule
POS & 91.95 & 87.06 & 95.55 \\ 
SEM & 82.00 & 81.11 & 91.41 \\
\bottomrule
\end{tabular}
\caption{POS and SEM tagging accuracy with baselines and an upper bound. MFT: most frequent tag; UnsupEmb: classifier using unsupervised word embeddings; Word2Tag: upper bound encoder-decoder. 
}
\label{tab:results-baselines}
\end{table}

Table~\ref{tab:results-baselines} shows baseline and upper bound results. 
The UnsupEmb baseline performs rather poorly on both POS and SEM tagging.
In comparison, NMT word embeddings (Table~\ref{tab:results-semtags-4layers}, rows with $k=0$) perform slightly better, suggesting that word embeddings learned as part of the NMT model are better syntactic and semantic representations. 
However, the results are still below the most frequent tag baseline (MFT), indicating that non-contextual word embeddings are poor representations for POS and SEM tags.

\subsection{Effect of network depth}

Table~\ref{tab:results-semtags-4layers}  summarizes the results of training classifiers to predict POS and SEM tags using features extracted from different encoding layers of 4-layered NMT systems.\footnote{The results given are with a unidirectional encoder; in section~\ref{sec:variants} we compare with a bidirectional encoder and observe similar trends.} 
In the POS tagging results (first block), 
as the representations move above layer 0, performance jumps to around 91--92\%. This is above the UnsupEmb baseline but only on par with the MFT baseline (Table~\ref{tab:results-baselines}). We note that previous work reported performance above a majority baseline for POS tagging \cite{shi-padhi-knight:2016:EMNLP2016,belinkov:2017:ACL},
but they used a weak global majority baseline whereas we compare with a stronger, most frequent tag baseline. 
The results are also far below the Word2Tag upper bound (Table \ref{tab:results-baselines}).

\begin{table}[t]
\centering
\begin{tabular}{l|lllll|l} 
\toprule
$k$ & \multicolumn{1}{c}{Ar} & \multicolumn{1}{c}{Es} & \multicolumn{1}{c}{Fr} & \multicolumn{1}{c}{Ru} & \multicolumn{1}{c}{Zh} & \multicolumn{1}{|c}{En} \\ 
\midrule
\multicolumn{7}{c}{POS Tagging Accuracy} \\
\midrule
0 & 88.0 & 87.9 & 87.9 & 87.8 & 87.7 & 87.4 \\ 
1 & 92.4 & 91.9 & 92.1 & 92.1 & 91.5 & 89.4 \\ 
2 & 91.9 & 91.8 & 91.8 & 91.8 & 91.3 & 88.3 \\ 
3 & 92.0 & 92.3 & 92.1 & 91.6 & 91.2 & 87.9 \\ 
4 & 92.1 & 92.4 & 92.5 & 92.0 & 90.5 & 86.9 \\ 
\midrule
\multicolumn{7}{c}{SEM Tagging Accuracy} \\
\midrule
0 & 81.9 & 81.9 & 81.8 & 81.8 & 81.8 & 81.2 \\ 
1 & 87.9 & 87.7 & 87.8 & 87.9 & 87.7 & 84.5 \\ 
2 & 87.4 & 87.5 & 87.4 & 87.3 & 87.2 & 83.2 \\ 
3 & 87.8 & 87.9 & 87.9 & 87.3 & 87.3 & 82.9 \\ 
4 & 88.3 & 88.6 & 88.4 & 88.1 & 87.7 & 82.1 \\ 
\midrule
\multicolumn{7}{c}{BLEU} \\
\midrule
& 32.7 & 49.1 & 38.5 & 34.2 & 32.1 & 96.6 \\ 
\bottomrule
\end{tabular}
\caption{SEM and POS tagging accuracy  
using features extracted from the $k$-th encoding layer  
of 4-layered NMT models  trained with different target languages. 
``En'' column is an English autoencoder.
BLEU scores are given for reference. 
}
\label{tab:results-semtags-4layers}
\end{table}

Comparing layers 1 through 4, we see that in 3/5 target languages (Ar, Ru, Zh), POS tagging accuracy peaks at layer 1 and does not improve at higher layers, with some drops at layers 2 and 3. In 2/5 cases (Es, Fr) the performance is higher at layer 4. This result is partially consistent with previous findings regarding the quality of lower layer representations for the POS tagging task \cite{shi-padhi-knight:2016:EMNLP2016,belinkov:2017:ACL}. 
One possible explanation for the discrepancy when using different target languages, is that French and Spanish are typologically closer to English compared to the other languages. It is possible that when the target language is similar to the source language, they both have similar POS characteristics, leading to more benefit in using upper layers for POS tagging. 

Turning to SEM tagging (Table~\ref{tab:results-semtags-4layers}, second block), 
representations from layers 1 through 4 boost the performance 
to around 87-88\%, which is far above the UnsupEmb and MFT baselines. While these results are below the oracle Word2Tag results (Table \ref{tab:results-baselines}), they indicate that NMT representations contain useful information for SEM tagging. 

Going beyond the 1st encoding layer, representations from the 2nd and 3rd layers  
do not consistently improve semantic tagging performance. However, representations from the 4th layer 
lead to significant improvement with all target languages except for Chinese. 
Note that there is a statistically significant difference ($p<0.001$) between layers 0 and 1 for all target languages, and between layers 1 and 4 for all languages except for Chinese, according to the approximate randomization test \cite{sigf06}.

Intuitively, higher layers have a more global perspective because they have access to higher representations of the word and its context, while lower layers have a more local perspective. Layer 1 has access to context but only through one hidden layer which may not be sufficient for capturing semantics.
It appears that higher representations are necessary for learning even relatively simple lexical semantics.

Finally, we  found that En-En encoder-decoders (that is, English autoencoders) produce poor representations for POS and SEM tagging (last column in Table \ref{tab:results-semtags-4layers}). This is especially true with higher layer representations (e.g.\ around 5\% below the MT models using representations from layer 4). In contrast, the autoencoder has 
 excellent sentence recreation capabilities (96.6 BLEU). This indicates that learning to translate (to any foreign language) is important for obtaining useful representations for both tagging tasks.

\subsection{Effect of target language}

Does translating into different languages make the  NMT system learn different  source-side representations?
Previous work reported a fairly consistent effect of the target language on the quality of NMT encoder representations for POS and morphological tagging~\cite{belinkov:2017:ACL}; they observed differences of $\sim$2-3\% in accuracy. We would like to examine if such an effect exists for both POS and SEM tagging.

\begin{table}[t]
\centering
\begin{tabular}{l|lllll|l} 
\toprule
 & \multicolumn{1}{c}{Ar} & \multicolumn{1}{c}{Es} & \multicolumn{1}{c}{Fr} & \multicolumn{1}{c}{Ru} & \multicolumn{1}{c}{Zh} & \multicolumn{1}{|c}{En} \\ 
\midrule
POS & 88.7 & 90.0 & 89.6 & 88.6 & 87.4 & 85.2 \\ 
\midrule
SEM & 85.3 & 86.1 & 85.8 & 85.2 & 85.0 & 80.7 \\ 
\bottomrule
\end{tabular}
\caption{SEM and POS tagging accuracy 
using features extracted from the 4th NMT encoding layer, trained with different target languages on a smaller parallel corpus (200K sentences).
}
\label{tab:results-semtags-4layers-small-data}
\end{table}

Table \ref{tab:results-semtags-4layers} 
also shows results using features obtained by training NMT systems on different target languages (the English source remains fixed). 
In both POS and SEM tagging,  
there are very small differences with different target languages ($\sim$0.5\%), except for Chinese which leads to slightly worse representations. 
While the differences are small, they are mostly statistically significant. For example, at layer 4, all the pairwise comparisons with different target languages are  statistically significant ($p<0.001$) in SEM tagging, and all except for two pairwise comparisons (Ar vs.\ Ru and Es vs.\ Fr) are significant in POS tagging. 

The effect of the target language is much smaller than that reported by ~\newcite{belinkov:2017:ACL} for POS and morphological tagging. 
This discrepancy can be attributed to the fact that our NMT systems are trained on much larger corpora than theirs (10x), so it is possible that some of the differences disappear when the NMT model is of better quality. To verify this, we trained systems using a smaller data size (200K sentences), comparable to the size used by~\newcite{belinkov:2017:ACL}. The results are shown in Table~\ref{tab:results-semtags-4layers-small-data}. In this case, we observe a variance in classifier accuracy of 1-2\%, based on target language, which is consistent with \newcite{belinkov:2017:ACL}. This is true for both POS and SEM tagging. 
The differences in POS tagging accuracy are statistically significant ($p < 0.001$) for all pairwise comparisons except for Ar vs.\ Ru; the differences in SEM tagging accuracy are significant for all 
comparisons except for Ru vs.\ Zh.

Finally, we note that training an English autoencoder on the smaller dataset results in much worse representations compared to MT models, for both POS and SEM tagging (Table~\ref{tab:results-semtags-4layers-small-data}, last column), consistent with the behavior we observed on the larger data (Table~\ref{tab:results-semtags-4layers}, last column).

\begin{figure}[t]
\includegraphics[width=\linewidth]{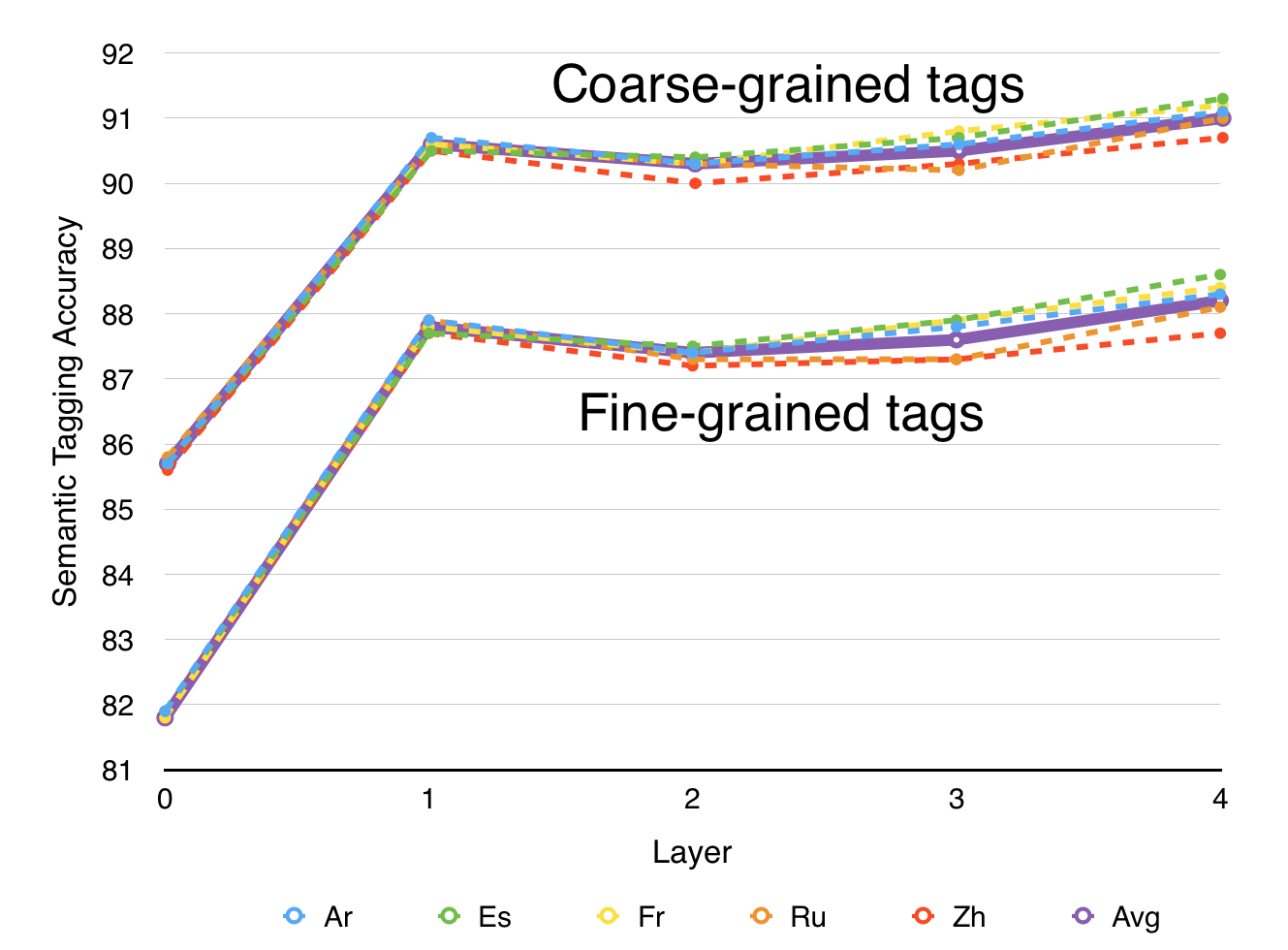}
\caption{SEM tagging accuracy with fine/coarse-grained tags using features extracted from different encoding layers 
of 4-layered NMT models  trained with different target languages. 
}
\label{fig:results-semtags-4layers}
\vspace{-10pt}
\end{figure}

\subsection{Analysis at the semantic tag level}
The SEM tags are grouped in coarse-grained categories such as events, names, time, and logical expressions \cite{bjerva-plank-bos:2016:COLING}.  
In Figure~\ref{fig:results-semtags-4layers} (top lines), 
we show the results of training and testing classifiers on coarse 
tags. Similar trends to the fine-grained case arise, with slightly higher absolute scores: significant improvement using the 1st encoding layer and some additional improvement using the 4th layer, both statistically significant ($p<0.001$). Again, there is a small effect of the target language.

\begin{figure}[t]
\includegraphics[width=\linewidth]{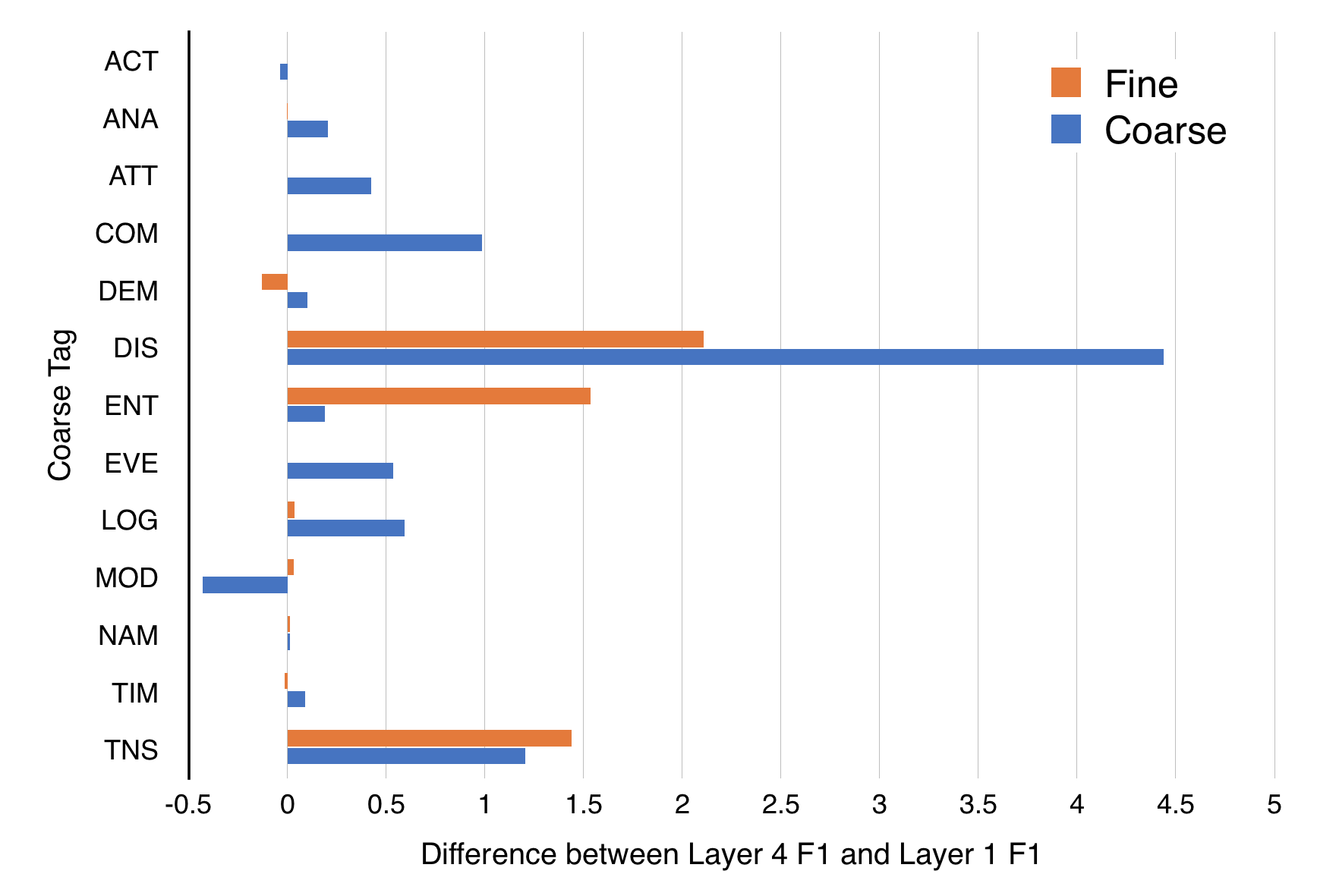}
\caption{Difference in F$_1$ when using representations from layer 4 compared to layer 1, showing F$_1$ when directly predicting coarse tags (blue) and when predicting fine-grained tags and averaging inside each coarse tag (red).}
\label{fig:coarse-layer1-4-f1}
\vspace{-15pt}
\end{figure}

Figure \ref{fig:coarse-layer1-4-f1} shows the change in F$_1$ score (averaged over target languages) when moving from layer 1 to layer 4 representations. The blue bars describe the differences per coarse tag when directly predicting 
coarse tags. The red bars show the same differences when predicting fine-grained tags and micro-averaging inside each coarse tag. The former shows the differences between the two layers at distinguishing among coarse tags. The latter gives an idea of the differences when distinguishing between 
fine-grained tags within a coarse category.
The first observation is that in the majority of cases there is an advantage for classifiers
trained with layer 4
representations, i.e., higher layer representations are better suited to learn the SEM tags, at both coarse and fine-grained levels.

A number of additional observations can be made. It appears that higher layers of the NMT model are 
better at capturing semantic information such as: \emph{discourse relations} (\texttt{DIS} tag: 
subordinate vs.\ coordinate vs.\ apposition 
relations), semantic properties of nouns (\emph{roles} vs.\ \emph{concepts}, within the \texttt{ENT} tag), \emph{events} and \emph{predicate tense} (\texttt{EVE} and \texttt{TNS} tags), \emph{logic relations} and \emph{quantifiers} (\texttt{LOG} tag: 
disjunction, conjunction, implication, 
existential, universal, 
etc.), and \emph{comparative constructions} (\texttt{COM} tag: 
equatives, comparatives, and superlatives).
These examples represent semantic concepts and relations that require a level of abstraction going 
beyond the lexeme or word form, and thus might be better represented in higher layers in the deep 
network.

\begin{figure*}[t]
\centering
\footnotesize
\begin{tabular}{l|cc|p{12.3cm}}
\toprule
& L1 & L4 & \\
\midrule
1 & REL & \emph{SUB} & Zimbabwe 's President 
Robert Mugabe 
has freed three men who were jailed for murder and sabotage \underline{\emph{as}} they battled South Africa 's anti-apartheid African National Congress in 1988 . \\
2 & REL & \emph{SUB} & The military says the battle erupted \underline{\emph{after}} gunmen fired on U.S. troops and Afghan police investigating a reported beating of a villager . \\
3 & IST & \emph{SUB} & Election authorities had previously told Haitian-born Dumarsais Simeus that he was not eligible to run \underline{\emph{because}} he holds U.S. citizenship . \\	
\midrule
4 & AND & \emph{COO} & Fifty people representing 26 countries took the Oath of Allegiance this week ( Thursday ) \underline{\emph{and}} became U.S. citizens in a special ceremony at the Newseum in Washington , D.C. \\
5 & AND & \emph{COO} & But rebel groups said on Sunday they would not sign \underline{\emph{and}} insisted on changes . \\
6 & AND & \emph{COO} & A Fox asked him , `` How can you pretend to prescribe for others , when you are unable to heal your own lame gait \underline{\emph{and}} wrinkled skin ? '' \\
\midrule
7 & NIL & \emph{APP} & But Syria 's president \underline{\emph{,}} Bashar al-Assad , has already rejected the commission 's request [...] \\ 
8 & NIL & \emph{APP} & Hassan Halemi \underline{\emph{,}} head of the pathology department at Kabul University where the autopsies were carried out , said hours of testing Saturday confirmed [...] \\ 
9 & NIL & \emph{APP} & Mr. Hu made the comments Tuesday during 
a meeting with Ichiro Ozawa \underline{\emph{,}} the leader of Japan 's main opposition party . \\
\midrule
10 & \emph{AND} & COO & [...] 
abortion opponents will march past the U.S. Capitol \underline{\emph{and}} end outside the Supreme Court . \\
11 & \emph{AND} & COO & Van Schalkwyk said no new coal-fired power stations would be approved unless they use technology that captures \underline{\emph{and}} stores carbon emissions . \\
12 & \emph{AND} & COO & A MEMBER of the Kansas Legislature meeting a Cake of Soap was passing it by 
without recognition , 
but the Cake of Soap insisted on stopping \underline{\emph{and}} shaking hands . \\
\bottomrule
\end{tabular}
\caption{Examples of cases of disagreement between layer 1 (L1) and layer 4 (L4) representations when predicting SEM tags. The correct tag is \emph{italicized} and the relevant word is  \underline{\emph{underlined}}.}
\label{fig:error-analysis}
\vspace{-5pt}
\end{figure*}

One negative example that stands out in Figure~\ref{fig:coarse-layer1-4-f1} is the prediction of the 
\texttt{MOD} tag, 
corresponding to \emph{modality} (necessity, possibility, and negation). It seems that such 
semantic concepts should be better represented in higher layers following our previous hypothesis. Still layer 1 is better than layer 4 in this case. One possible explanation is that 
words  tagged as \texttt{MOD} form a closed class, 
with only a few and mostly unambiguous words (``no'', ``not'', ``should'', ``must'', ``may'', ``can'', ``might'', 
etc.). 
It is enough for the classifier to memorize these 
words in order to predict  this class with high F$_1$, and this is something that occurs better in lower layers. One final case worth mentioning is the \texttt{NAM} category, which stands for different types of named entities (person, location, organization, artifact, 
etc.). In principle, this seems a clear case of 
semantic abstractions  suited for higher layers, but the results from layer 4 are not significantly better than those from layer 1. This might be signaling a limitation of the NMT system at learning this type of semantic classes. Another factor might be the fact that many named entities are out of vocabulary words for the NMT system.

\subsection{Analyzing discourse relations}

In this section, we analyze specific cases of disagreement between predictions using representations from layer 1 and layer 4. We focus on discourse relations, as they show the largest improvement when going from layer 1 to layer 4 representations (\texttt{DIS} category in Figure~\ref{fig:coarse-layer1-4-f1}). 
Intuitively, identifying discourse relations requires a relatively large context so it is expected that higher layers would perform better in this case. 

There are three discourse relations in the SEM tags annotation scheme: subordinate (\texttt{SUB}), coordinate (\texttt{COO}), and apposition (\texttt{APP}) relations. For each of those, Figure~\ref{fig:error-analysis} (examples 1-9) shows the first three cases in the test set where layer 4 representations correctly predicted the tag but layer 1 representations were wrong. Examples 1-3 have subordinate conjunctions (\emph{as}, \emph{after}, \emph{because}) connecting a main and an embedded clause, which layer 4 is able to correctly predict. Layer 1 mistakes these as attribute tags (\texttt{REL}, \texttt{IST}) that are usually used for prepositions. In examples 4-5, the coordinate conjunction \emph{and} is used to connect sentences/clauses, which layer 4 correctly tags as \texttt{COO}. Layer 1 wrongly predicts the tag \texttt{AND}, which is used for conjunctions connecting shorter expressions like words (e.g.\ ``murder \emph{and} sabotage'' in example 1). Example 6 is probably an annotation error, as \emph{and} connects the phrases ``lame gait'' and ``wrinkled skin'' and should be tagged as \texttt{AND}. In this case, layer 1 is actually correct. 
In examples 7-9, layer 4 correctly identifies the comma as introducing an apposition, while layer 1 predicts \texttt{NIL}, a tag for punctuation marks without semantic content (e.g. end-of-sentence period).  
As expected, in most of these cases identifying the discourse function requires a fairly large context.

Finally, we show in examples 10-12 the first three occurrences of \texttt{AND} in the test set, where layer 1 was correct and layer 4 was wrong. Interestingly, two of these (10-11) are clear cases of \emph{and} connecting clauses or sentences, which should have been annotated as \texttt{COO}, and the last (12) is a conjunction of two gerunds. The predictions from layer 4 in these cases thus appear justifiable.

\vspace{-5pt}
\subsection{Other architectural variants}
\label{sec:variants}
Here we consider two architectural variants that have been shown to benefit NMT systems: 
bidirectional encoder and residual connections. 
We also experiment with NMT systems trained with different depths. 
Our motivation in this section is to see if the patterns we observed thus far hold in different NMT architectures. 

\paragraph{Bidirectional encoder}
Bidirectional LSTMs have become ubiquitous in NLP 
and also give some improvement as NMT encoders \cite{britz2017massive}. 
We confirm these results and note improvements in both translation (+1-2 BLEU) and SEM tagging quality (+3-4\% accuracy), across the board, when using a bidirectional encoder. 
Some of our bidirectional models obtain 92-93\% accuracy, which is close to the state-of-the-art on this task \cite{bjerva-plank-bos:2016:COLING}. 
We observed similar improvements on POS tagging. 
Comparing POS and SEM tagging (Table \ref{tab:results-uni-bi}), we note 
that higher layer representations improve SEM tagging, while POS tagging peaks at layer 1, in line with our previous observations.

\paragraph{Residual connections} Deep networks can sometimes be trained better if residual connections are introduced between layers. Such connections were also found useful for SEM tagging \cite{bjerva-plank-bos:2016:COLING}. Indeed, we noticed  small but consistent improvements in both translation  (+0.9 BLEU) and  POS and SEM tagging  (up to +0.6\% accuracy) when using features extracted from an NMT model trained with residual connections (Table~\ref{tab:results-uni-bi}). We also observe similar trends as before: POS tagging does not benefit from features from the upper layers, while SEM tagging improves with layer 4 representations.

\paragraph{Shallower MT models}
In comparing network depth in NMT, 
\newcite{britz2017massive} found that encoders with 2 to 4 layers performed the best. 
For completeness, we report here results using features extracted from models trained originally with 2 and 3 layers, in addition to our basic setting of 4 layers. Table~\ref{tab:results-2-3-4} shows consistent trends with our previous observations: POS tagging does not benefit from upper layers, while SEM tagging does, although the improvement is rather small in the shallower models.

\section{Related Work}
\label{sec:related-work}
Techniques for analyzing neural network models include visualization of hidden units~\cite{elman1991distributed,karpathy2015visualizing,kadar2016representation,qian-qiu-huang:2016:EMNLP2016}, which provide illuminating, but often anecdotal information on how the network works. A number of  studies aim to obtain quantitative correlations between parts of the neural network and linguistic properties, in both
speech~\cite{wang2017gate,wu2016investigating,alishahi2017encoding} and 
language processing models~\cite{kohn:2015:EMNLP,qian-qiu-huang:2016:EMNLP2016,qian-qiu-huang:2016:P16-11,adi2016fine,linzen2016assessing}. 
Methodologically, our work is most similar to ~\newcite{shi-padhi-knight:2016:EMNLP2016} and \newcite{belinkov:2017:ACL}, who also used hidden vectors from neural MT models to predict linguistic properties.  However, they focused on relatively low-level tasks (syntax and morphology, respectively), while we apply the approach to a semantic task and compare the results with a POS tagging task.

\begin{table}[t]
\centering
\begin{tabular}{l|l|ccccc}
\toprule
& & 0 & 1 & 2 & 3 & 4 \\
\midrule
\multirow{2}{*}{Uni} & POS & 87.9 & 92.0 & 91.7 & 91.8 & 91.9 \\
& SEM & 81.8 & 87.8 & 87.4 & 87.6 & 88.2 \\
\midrule
\multirow{2}{*}{Bi} & POS & 87.9 & 93.3 & 92.9 & 93.2 & 92.8 \\
& SEM & 81.9 & 91.3 & 90.8 & 91.9 & 91.9 \\
\midrule
\multirow{2}{*}{Res} & POS & 87.9 & 92.5 & 91.9 & 92.0 & 92.4 \\
& SEM & 81.9 & 88.2 & 87.5 & 87.6 & 88.5 \\
\bottomrule
\end{tabular}
\caption{POS and SEM tagging accuracy with features from different layers of 4-layer 
Uni/Bidirectional/Residual 
NMT encoders, averaged over all 
non-English target languages.}
\label{tab:results-uni-bi}
\end{table}

\begin{table}[t]
\centering
\begin{tabular}{l|l|ccccc}
\toprule
& & 0 & 1 & 2 & 3 & 4 \\
\midrule
\multirow{2}{*}{4} & POS & 87.9 & 92.0 & 91.7 & 91.8 & 91.9 \\
& SEM & 81.8 & 87.8 & 87.4 & 87.6 & 88.2 \\
\midrule
\multirow{2}{*}{3} & POS & 87.9 & 92.5 & 92.3 & 92.4 & --  \\
& SEM & 81.9 & 88.2 & 88.0 & 88.4 & -- \\
\midrule
\multirow{2}{*}{2} & POS & 87.9 & 92.7 & 92.7 & -- & -- \\
& SEM & 82.0 & 88.5 & 88.7 & -- & -- \\
\bottomrule
\end{tabular}
\caption{POS and SEM tagging accuracy with features from different layers of 
2/3/4-layer encoders, averaged over all 
non-English target languages.}
\label{tab:results-2-3-4}
\end{table}

\section{Conclusion}
\label{sec:conclusion}

While neural network  models  
have improved the state-of-the-art in machine translation,  
it is difficult to interpret what they learn about language. In this work, we explore what kind of linguistic information such models learn at different layers.  
Our experimental evaluation leads to interesting insights about the hidden representations in NMT models such as the effect of layer depth and target language on part-of-speech and semantic tagging. 

In the future, we would like to extend this work to other syntactic and semantic tasks that require building relations such as  
dependency relations or predicate-argument structure. We believe that understanding how semantic properties are learned 
in NMT is a key step for creating better MT systems. 

\section*{Acknowledgments}
This research was carried out in collaboration between the HBKU Qatar Computing Research Institute (QCRI) and the MIT Computer Science and Artificial Intelligence Laboratory (CSAIL).

\bibliography{emnlp2017}
\bibliographystyle{ijcnlp2017}

\end{document}